\documentclass[letterpaper]{article} 
\usepackage{aaai24}  
\usepackage{times}  
\usepackage{helvet}  
\usepackage{courier}  
\usepackage[hyphens]{url}  
\usepackage{graphicx} 
\usepackage{natbib}  
\usepackage{caption} 
\usepackage{algorithm}
\usepackage{algorithmic}
\usepackage{graphicx}
\usepackage{amsmath}
\usepackage{amssymb}
\usepackage{booktabs}
\usepackage{xcolor,colortbl}
\usepackage{color, colortbl}
\definecolor{tabhighlight}{HTML}{e5e5e5}
\definecolor{citecolor}{HTML}{0071bc}
\newcommand{\tableCellHeight}{1}
\newcommand{\tabstyle}[1]{
  \setlength{\tabcolsep}{#1}
  \renewcommand{\arraystretch}{\tableCellHeight}
  \centering
  \small
}
\usepackage{subcaption}
\usepackage{sidecap}
\newcommand{\tablestyle}[2]{\setlength{\tabcolsep}{#1}\renewcommand{\arraystretch}{#2}\centering\footnotesize}
\usepackage{newfloat}
\usepackage{listings}
\usepackage[capitalize]{cleveref}
\crefname{section}{Sec.}{Secs.}
\Crefname{section}{Section}{Sections}
\Crefname{table}{Table}{Tables}
\crefname{table}{Tab.}{Tabs.}
\DeclareCaptionStyle{ruled}{labelfont=normalfont,labelsep=colon,strut=off} 
\floatstyle{ruled}
\newfloat{listing}{tb}{lst}{}
\floatname{listing}{Listing}
\title{Domain-Controlled Prompt Learning}
\author{
 Qinglong Cao\textsuperscript{\rm 1,2 }, Zhengqin Xu\textsuperscript{\rm 1}, Yuntian Chen\textsuperscript{\rm 2}\thanks{ Yuntian Chen is the corresponding author}, Chao Ma\textsuperscript{\rm 1}, Xiaokang Yang\textsuperscript{\rm 1}
}
\affiliations{
    \textsuperscript{\rm 1}MoE Key Lab of Artificial Intelligence, AI Institute, Shanghai Jiao Tong University, Shanghai, China\\
    \textsuperscript{\rm 2}Ningbo Institute of Digital Twin, Eastern Institute of Technology, Ningbo, China\\
 \{caoql2022, fate311\}@sjtu.edu.cn, ychen@eitech.edu.cn, \{chaoma, xkyang\}@sjtu.edu.cn\\
}

\usepackage{bibentry}
\makeatletter
\def\@copyrightspace{\relax}
\makeatother

\begin{document}
\maketitle

\begin{abstract}
Large pre-trained vision-language models, such as CLIP, have shown remarkable generalization capabilities across various tasks when appropriate text prompts are provided. However, adapting these models to specific domains, like remote sensing images (RSIs), medical images, etc, remains unexplored and challenging. Existing prompt learning methods often lack domain-awareness or domain-transfer mechanisms, leading to suboptimal performance due to the misinterpretation of specific images in natural image patterns.  To tackle this dilemma, we proposed a \textbf{Domain-Controlled Prompt Learning} for the specific domains. Specifically, the large-scale specific domain foundation model (LSDM) is first introduced to provide essential specific domain knowledge. Using lightweight neural networks, we transfer this knowledge into domain biases, which control both the visual and language branches to obtain domain-adaptive prompts in a directly incorporating manner.  Simultaneously, to overcome the existing overfitting challenge, we propose a novel noisy-adding strategy, without extra trainable parameters, to help the model escape the suboptimal solution in a global domain oscillation manner. Experimental results show our method achieves state-of-the-art performance in specific domain image recognition datasets. Our code is available at \url{https://github.com/caoql98/DCPL.}

\end{abstract}

\section{Introduction}
\label{sec:intro}
With the emergence of deep learning technology, various visual understanding tasks, including classification~\cite{simonyan2014very,he2016deep}, semantic segmentation~\cite{long2015fully,chen2017deeplab}, and object detection~\cite{redmon2016you,girshick2015fast}, have witnessed remarkable progress. However, the success of these tasks heavily relies on access to large-scale, high-quality annotated datasets~\cite{deng2009imagenet,lin2014microsoft}, which entail significant labor and expense for each specific visual task. To tackle this practical challenge, the Contrastive Language-Image Pretraining (CLIP)~\cite{radford2021learning} has been introduced, aiming to provide transferable visual features that can be leveraged across a diverse range of downstream tasks. By employing contrastive learning with extensive image-text pairs, CLIP has demonstrated exceptional zero-shot generalization capabilities.

\begin{figure}[t]
	\begin{center}
		\includegraphics[width=0.92\linewidth]{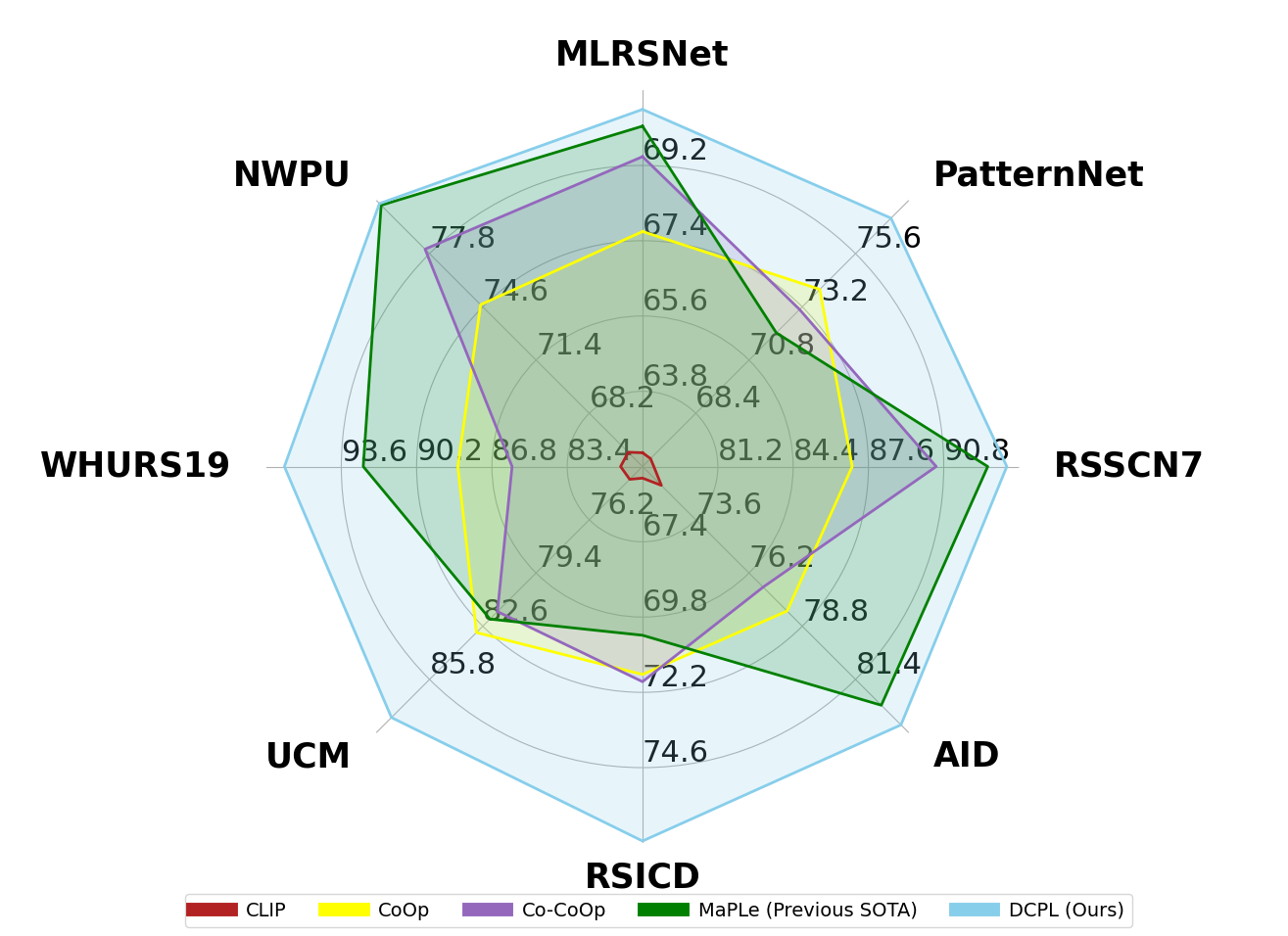}
	\end{center}
  \vspace{-0.4cm}
	\caption{Using RSIs as examples. Our method achieves state-of-the-art performance on 8 different RSIs datasets in harmonic mean (HM) for novel class generalization task.    }
	\label{fig:first}
 \vspace{-0.9cm}
\end{figure}

In CLIP, visual categories are directly incorporated into carefully designed templates as prompts. Nonetheless, the creation of appropriate templates can be a time-consuming endeavor. To address this concern and drawing inspiration from prompt learning techniques, CoOp~\cite{zhou2022learning} proposes context optimization, employing learnable context vectors to enhance the zero-shot generalization performance. Following the prompt learning paradigm, numerous prompt learning algorithms have been developed for vision-language models, yielding notable advancements in zero-shot image recognition. For instance, CoCoOp~\cite{zhou2022conditional} tackles the class shift problem by introducing input-conditional tokens, while MaPLe adopts prompt learning for both vision and language branches to enhance the alignment between visual and linguistic representations.

Despite the progress made in prompt learning algorithms~\cite{zhou2022learning,zhou2022conditional,khattak2023maple,wang2022learning,ge2022domain}, they only consider the same-domain downstream task, while the adaptation problem from the natural image domain to specific domains like  RSIs has rarely been considered.  The domain-awareness or domain-transfer mechanisms are correspondingly been ignored. Naturally, existing prompt learning algorithms would approach these domain-specific images with inappropriate natural image perception patterns, leading to suboptimal performance in specific domain recognition tasks.

To address this challenge and enable prompt learning to effectively model the necessary domain adaptations for specific domains like RSIs, medical images, etc, we propose a novel domain-controlled prompt learning approach. Our key idea is to generate domain-adaptive prompts for both the visual and language branches, instead of relying on existing domain-insensitive prompts, and experiments are implemented on RSIs and medical images to demonstrate the efficiency. Specifically, we introduce the newly open-sourced large-scale specific domain foundation model (LSDM) as the specific domain knowledge. By incorporating lightweight neural networks, the LSDM generates domain biases separately for the visual branch and the language branch. The domain bias for language is incorporated into the learnable context vector, while the domain bias for the visual branch is directly integrated into the image features. This approach controls the model to rightly understand the specific domain data, leading to more informed and contextually rich representations, ultimately enhancing the model's discriminative power and overall performance.

Meanwhile, CoCoOp has identified  the overfitting problem  caused by category shift from base to novel category and attempted to solve it in a conditional method.  However, we tend to tackle it in a more explicit manner. To straightforwardly solve this,  adopting  dropout or mutation operations seems to be  a plausible solution.  However, these strategies only  introduce randomness and variations to some extent, they are still constrained by their local-sampling nature (dropout) and point-based modifications (mutation), which means they are insufficient for  escaping suboptimal solutions.

Inspired by the random sampling process in diffusion models~\cite{ho2020denoising,nichol2021improved}, which greatly facilitates exploration in complex spaces, we propose a novel noisy-adding strategy to handle it. This strategy induces global domain oscillation throughout the entire feature space by introducing adaptive random Gaussian noise. In contrast to local sampling and point jittering, this strategy allows for broader exploration across  whole feature space, preventing model from being trapped in narrow solution regions. As shown in Figure~\ref{fig:first}, our proposed method outperforms existing prompt learning approaches across 8 diverse remote sensing image recognition datasets.

To sum up, the main contributions of our proposed domain-controlled prompt learning could be concluded as follows:
\begin{itemize}
	\item To the best of our knowledge, we propose the first prompt learning paradigm for specific domains. By introducing the large-scale specific domain foundation model (LSDM), the proposed domain-controlled prompt learning provides better domain-adaptive prompts for both the visual and language branches.
	
	\item A novel noise-adding strategy is proposed to explicitly address the issue of overfitting in domain-controlled prompt learning. This strategy enables the model to avoid getting stuck in suboptimal solutions and instead explore a wider solution space in search of the best solution.

	\item Taking RSIs and medical images for instance. Our method is evaluated on  specific domain datasets and extensive ablation studies are conducted to testify its characteristics. Due to space limitations, more detailed domain experiments  are illustrated in supplementary materials. The experiments of various special tasks demonstrate our method achieves state-of-the-art performance.
	
\end{itemize}

 \begin{figure*}[ht]
	\begin{center}
		\includegraphics[width=0.85\linewidth]{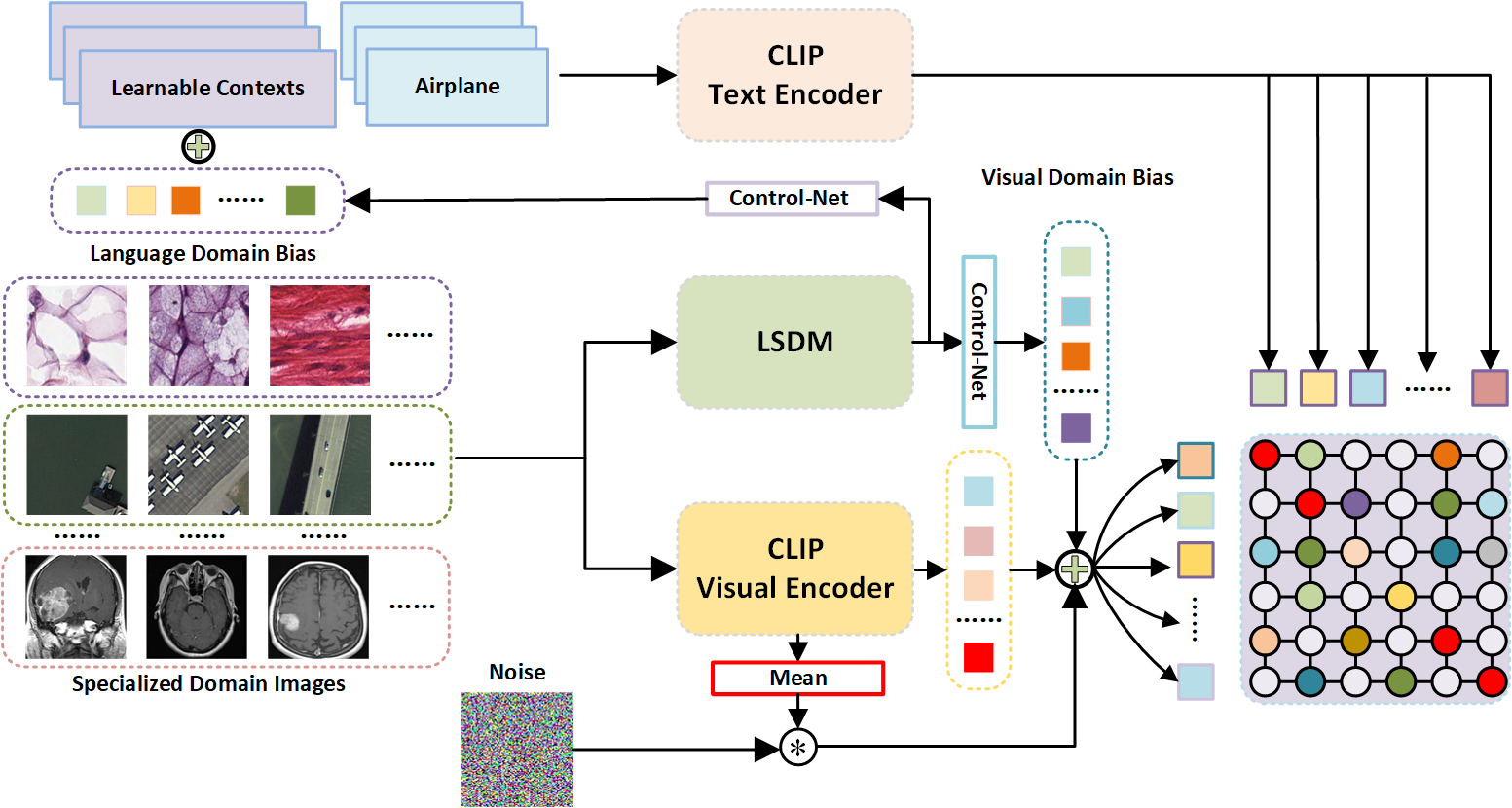}
	\end{center}
  	  \vspace{-0.4cm}
	\caption{Overview of our proposed Domain-Controlled Prompt Learning~(DCPL) framework.  Introducing the large-scale specific domain foundation model~(LSDM) to provide domain foundation knowledge, DCPL provides domain-adaptive prompts to respectively control the visual and language branch in a directly incorporating manner. Additionally, a noisy-adding strategy is further proposed to help the model escape the suboptimal solution in a global domain oscillation manner. }
	\label{fig:second}
  \vspace{-0.6cm}
\end{figure*}

\section{Related Work}

\textbf{Vision Language Models.} Vision Language (V-L) models aim to build a cohesive alignment between images and languages  to learn a shared embedding space that encompasses both modalities. Conventional V-L models typically comprise three key components: a visual encoder, a text encoder, and an alignment loss.  Visual images were often processed using hand-crafted descriptors~\cite{elhoseiny2013write,socher2013zero} or neural networks~\cite{frome2013devise,lei2015predicting}, while texts were typically encoded using pre-trained word vectors~\cite{socher2013zero,frome2013devise} or frequency-based descriptors~\cite{schnabel2015evaluation,gong2018frage}. The visual and textual representations were then aligned using techniques like metric learning~\cite{frome2013devise} or multi-label classification~\cite{joulin2016learning,gomez2017self}.

However, recent advancements in V-L models~\cite{radford2021learning,jia2021scaling,yao2021filip,yuan2021florence,zhai2022lit} have revolutionized the field by seamlessly integrating the two modalities through joint learning of image and text encoders in an image-text pair alignment fashion. For example, models like CLIP~\cite{radford2021learning} and ALIGN~\cite{jia2021scaling} leverage an extensive corpus of approximately 400 million and 1 billion image-text pairs, respectively, to train their multi-modal networks. This approach enables the recent V-L models to generate highly informative cross-modality representations, leading to exceptional performance across various downstream tasks, including few-shot and zero-shot visual recognition~\cite{gao2021clip,zhang2021tip}. Furthermore, by carefully tailoring V-L models and effectively utilizing the cross-modality representations, traditional image recognition~\cite{conde2021clip,fu2022cma}, object detection~\cite{feng2022promptdet,bangalath2022bridging}, and semantic segmentation~\cite{li2022language,luddecke2022image,rao2022denseclip} tasks have also achieved promising performance improvements.

\textbf{Prompt Learning in Vision Language models.}
V-L models can be adapted to downstream tasks using either full fine-tuning or linear probing approaches. However, full fine-tuning is computationally intensive and may degrade the previously learned cross-modality representations. On the other hand, linear probing limits the zero-shot capability of models like CLIP. To address these challenges, inspired by prompt learning in natural language processing, many algorithms~\cite{zhou2022learning,zhou2022conditional,khattak2023maple,wang2022learning,ge2022domain} have been  proposed to efficiently adapt V-L models in the prompt tokens learning manner. For instance,  CoOp~\cite{zhou2022learning} introduces context optimization to adapt CLIP by using learnable context vectors while keeping the pre-trained parameters fixed. However, CoOp's learned context has limited generalizability and suffers from overfitting issues in base categories. To overcome these limitations, CoCoOp~\cite{zhou2022conditional} proposes conditional context optimization, which provides instance-conditioned prompt tokens. While previous methods focus on efficient prompt learning in CLIP's language branch, the visual branch is few considered. Addressing this gap, MaPLe~\cite{khattak2023maple} proposes multi-modal prompt learning to simultaneously adapt vision and language representations, which successfully improves the  cross-alignment. However, existing algorithms rarely consider the adaptation problem when transitioning from the natural image domain to specific domains like RSIs, and medical images. This lack of domain awareness or domain-transfer mechanism leads to an inadequate perception of specific domain images and results in suboptimal performance. To handle this issue, we propose domain-controlled prompt learning for specific domain images to provide domain-adaptive prompts.

\section{Method}
To provide domain-adaptive prompts, our proposed method (DCPL) first introduces the large-scale remote sensing foundation model into CLIP to achieve domain-controlled prompt learning. Figure~\ref{fig:second} shows the overall architecture of the proposed network. More specifically, unlike previous methods, to better transfer CLIP from the natural domain to specific domains, the large-scale specific domain foundation model (LSDM) is first introduced to provide specific domain features as the specific domain knowledge. Then, through the designed control nets, the specific domain features could be respectively transferred into language domain bias and visual domain bias. By adding the language domain bias into the learnable context vectors and incorporating the visual domain bias into the visual features, the networks are controlled to have domain-adaptive prompts. Simultaneously, to help the network escape the suboptimal solution and search solution in a broader space, the noise is adaptively added to the visual features to perform a global domain oscillation. Below we first introduce the pre-trained CLIP~\cite{radford2021learning} and the introduced LSDM~\cite{wang2022advancing, ma2023segment}. Then, we illustrate the proposed DCPL.
\subsection{Review of CLIP}
CLIP mainly contains a visual encoder and a text encoder, which could respectively generate image embeddings and corresponding text embeddings. We follow the setting in previous methods~\cite{khattak2023maple, zhou2022conditional} to adopt the vision transformer (ViT~\cite{dosovitskiy2020image}) based CLIP model.

For the visual encoder, $I \in {\mathbb{R}^{H \times W \times 3}}$ would be firstly spilled into M fixed-size patches, which are further reshaped as patch embeddings ${E_p} \in {\mathbb{R}^{M \times {d_p}}}$. Then the patch embeddings would be propagated into the transformer layers with the learnable category tokens $C_p$. To obtain the final image embeddings $x \in {\mathbb{R}^{d_t}}$, the category tokens $C_l$ from last layer would be projected into the common Visual-Language feature space:
\begin{equation}
x = VisProj({C_l})
\end{equation}

In the test encoder, the text descriptions for images would be first tokenized into the
words and further projecting them to word embedding $W_t$. Subsequently, the word embeddings would be inputted into transformer layers. Similarly, the text embeddings $W_l$ from the last layer are projected into the common Visual-Language feature space to obtain the final text embeddings $\omega \in {\mathbb{R}^{d_t}}$:
\begin{equation}
\omega = WordProj({w_l})
\end{equation}

With these image embeddings and the corresponding text embeddings, the CLIP would maximize the cosine similarity between the image and its matched text while minimize the cosine similarity between the image and its unmatched text. After training, the CLIP would be directly leveraged to perform the zero-shot classification. Particularly, the $\omega_i$ is generated from the hand-craft prompt, such as ``a photo of $category$'', where $category$ is the $i$-th class name. Then, suppose there are $C$ categories and the visual embedding of the image is $x$, the probability of the image belonging to $i$-th class name is produced by:
\begin{equation}
\label{eq3}
p(y|x) = \frac{{\exp (sim(x,\omega )/\tau )}}{{\sum\nolimits_{i = 1}^C {\exp (sim(x,\omega )} /\tau )}}
\end{equation}
where $sim$ denotes the cosine similarity and $\tau$ is the adjusting temperature parameter.

\subsection{Large-Scale Specific Domain Foundation Model }
The Large-Scale Specific Domain Foundation Model (LSDM)~\cite{sun2022ringmo,wang2022advancing, ma2023segment} is recently proposed to provide better representations for downstream image processing tasks like RSIs, medical images, etc. Inspired by this,  the large-scale remote sensing foundation model (LRSM) ~\cite{wang2022advancing} and MedSAM~\cite{ma2023segment} are respectively utilized to provide basic specific domain knowledge for RSIs and medical images. The LRSM mainly adopt ViT~\cite{dosovitskiy2020image} and ViTAE~\cite{xu2021vitae} architectures, and the networks are trained in an MAE~\cite{he2022masked} manner with millions of RSIs.  MAE aims to recover the masked images with the visible parts in an encoder-decoder architecture. The network is optimized by minimizing the loss between the recovered regions and the ground-truth masked regions.  Harnessing the power of a meticulously curated dataset comprising over one million medical images, MedSAM~\cite{ma2023segment} are pre-trained for downstream medical image processing tasks. Since we need to control the visual and language branches in the cross-modality space, the pre-trained encoder of the LSDM network is leveraged to provide the specific domain embeddings  $R_b$ as the basic specific domain knowledge.

\subsection{DCPL: Domain-Controlled Prompt Learning }
Existing prompt learning methods all ignore the adaption problem from the natural domain to the specific domain like the remote sensing domain. This negligence would result in the specific domain images being handled with an inappropriate natural image processing pattern, further leading to suboptimal performance. To tackle this challenge, we introduce the LSDM to provide specific domain knowledge to control the visual and language to perceive the specific domain images  with domain-adaptive prompts.

Give the input images $I \in {\mathbb{R}^{H \times W \times 3}}$, the input images would be propagated into the pre-trained encoder of LSDM to generate the specific domain embeddings $R_b \in {\mathbb{R}^{d_r}}$ as the basic specific domain knowledge:
\begin{equation}
{R_b} = \mathop {Encoder}\limits_{LSDM} (I)
\end{equation}

\textbf{Control the Language Branch.} To control the language branch, we first adopt the learnable context vectors in the CoOp~\cite{zhou2022learning} as the basic prompt. Assuming we have $M$ context tokens $ \{ v_1^{ct},v_2^{ct},...,v_M^{ct}\}$. The language domain bias $D_b^{l} \in {\mathbb{R}^{d_t}}$ for the language branch is generated by  $D_b^{l} = {f_{LC}}( R_b)$, where  ${f_{LC}}( \cdot )$ denotes the designed language control net. By directly incorporating the domain bias $D_b$ into the context tokens, the basic context tokens are transferred into the specific domain:
\begin{equation}
v_m^{ct}({R_b}) = v_m^{ct} + D_b^{l},m \in \{ 1,2,...M\}  
\end{equation}
Then the final domain-adaptive prompt could be defined as  $ {t_i}({R_b}) = \{ v_1^{ct}({R_b}),v_2^{ct}({R_b}),...,v_M^{ct}({R_b}),{C_i}\} $, where $i$ denotes $i$-th category, and $C_i$ means the category name.

\textbf{Control the Visual Branch.} We first compute the visual domain bias $D_b^{v} \in {\mathbb{R}^{d_t}}$ through the designed visual control net${f_{LC}}( \cdot )$: $D_b^{v} = {f_{VC}}( R_b)$. Since the specific domain embeddings have the same modality as the final image embeddings $x \in {\mathbb{R}^{d_t}}$. Thus, the generated domain bias could be directly fused with  $x$ to directly generate the domain-adaptive image features ${x_d}({R_b})$:
\begin{equation}
{x_d}({R_b}) = x + D_b^{v}
\end{equation}
In this manner, both prompts for the visual and language branches are directly controlled by the introduced specific domain knowledge $R_b$, which helps the model process the specific domain images in a correct specific domain perception manner.

\textbf{Noisy-Adding Strategy.} As discussed in CoCoOp~\cite{zhou2022conditional}, prompt learning methods tend to be overfitted in the base categories and  not generalizable to wider unseen classes within the same task. CoCoOp tends to handle this problem with an instance-conditional network. However, we tend to solve it in a more explicit manner. Normally, we could adopt the dropout or mutation operations to solve this. However, these methods are actually local sampling strategies or point-based modifications. This means the prompt learning network is still searching for solutions in the oscillation-limited domain. 

The inference process in diffusion models ~\cite{ho2020denoising,nichol2021improved} would  add random noise to escape the trivial solutions and search for better solutions in complex space. Inspired by this,  we also randomly sample the noise to help the model escape the suboptimal solution and search for the solutions with the global domain oscillate. Particularly, given the Gaussian noise $z$, we first compute the adaptive adjusting factor $\sigma _m$ by computing the mean  of image embeddings: ${\sigma _m} = Mean(x)$. Then, the adaptive adjusting factor is leveraged to scale the sampled Gaussian noise. To directly handle the overfitting problem, the noise would be directly added to the domain-adaptive images features ${x_d}({R_b})$:
\begin{equation}
{\widehat x_d}({R_b}) = {x_d}({R_b}) + {\sigma _m}z
\end{equation}
Finally, the probability of the image belonging to $i-$th category name is changed from equation~\ref{eq3} to:
\begin{equation}
p(y|x) = \frac{{\exp (sim({{\widehat x}_d}({R_b}),{t_y}({R_b}))/\tau )}}{{\sum\nolimits_{i = 1}^C {\exp (sim({{\widehat x}_d}({R_b}),{t_i}({R_b}))/\tau )} }}
\end{equation}

\begin{table*}[t!]
\small
\tablestyle{6pt}{0}
\addtolength{\tabcolsep}{-6pt}
    \tabstyle{1.0pt}
    \setlength{\tabcolsep}{3.2pt}
    \caption{Comparison with existing methods in base-to-novel generalization on 8 remote sensing recognition datasets. The best results are shown in bold.
    }
     \vspace{-0.35cm}
    \scalebox{0.8}{
    \begin{subtable}[t]{.32\textwidth}
    \centering
    \caption{\textbf{Average over 8 datasets}}
    \begin{tabular}{l cc c}
    \toprule
    & Base & Novel & HM \\
    \midrule
    CLIP & 71.19	&71.33	&70.63 \\
    CoOp & 87.61	&70.84	&78.03 \\
    Co-CoOp & 91.82	&68.98	&78.43 \\
    MaPLe & 93.12	&71.71	&80.42 \\
    \midrule
    Ours (ViTAE) &93.07	&73.79	&81.81\\
    \rowcolor{tabhighlight}
       Ours (ViT) &  \textbf{93.77}	&\textbf{75.81}	&\textbf{83.36}\\
    \bottomrule
    \end{tabular}
    \end{subtable}
    }
        \scalebox{0.8}{
    \begin{subtable}[t]{.32\textwidth}
    \centering
    \caption{MLRSNet}
    \begin{tabular}{l cc c}
    \toprule
    & Base & Novel & HM \\
    \midrule
    CLIP & 64.50 & \textbf{60.30} & 62.33 \\
    CoOp & 79.37 & 58.90 & 67.62\\
    Co-CoOp & 83.30 & 59.50 & 69.42 \\
    MaPLe & 85.23 & 59.60 & 70.15 \\
    \midrule
    Ours (ViTAE) & 86.30 & 58.47 & 69.71 \\
    \rowcolor{tabhighlight}
     Ours (ViT) &  \textbf{87.05} &  59.30 & \textbf{70.54} \\
    \bottomrule
    \end{tabular}
    \end{subtable}
    }
    ~
        \scalebox{0.8}{
    \begin{subtable}[t]{.32\textwidth}
    \centering
    \caption{PatternNet}
    \begin{tabular}{l cc c}
    \toprule
    & Base & Novel & HM \\
    \midrule
    CLIP & 70.60 & 62.60 & 66.36 \\
    CoOp & 87.30 & 64.20 & 73.99 \\
    Co-CoOp & 93.70 & 59.90 & 73.08 \\
    MaPLe & 95.30 & 57.90 & 72.03 \\
    \midrule
    Ours (ViTAE) &  95.33 &62.07	&75.19 \\
    \rowcolor{tabhighlight}
     Ours (ViT) & \textbf{95.93} &	\textbf{64.60}	&\textbf{77.21}
 \\
    \bottomrule
    \end{tabular}
    \end{subtable}
    }
    ~
        \scalebox{0.8}{
    \begin{subtable}[t]{.32\textwidth}
    \centering
    \caption{RSSCN7}
    \begin{tabular}{l cc c}
    \toprule
    & Base & Novel & HM \\
    \midrule
    CLIP & 66.70 & 95.30 & 78.48 \\
    CoOp & 84.80  &	89.13 & 86.91 \\
    Co-CoOp & 90.97	&90.00	&90.48 \\
    MaPLe & 91.67	&93.70	&92.67 \\
    \midrule

    Ours (ViTAE) & 87.87	&92.13	&89.95\\
    \rowcolor{tabhighlight}
    Ours (ViT) &  \textbf{91.67}	&\textbf{95.37}	&\textbf{93.48}\\
    \bottomrule
    \end{tabular}
    \end{subtable}
        }
        \scalebox{0.8}{
    \begin{subtable}[t]{.32\textwidth}
    \centering
    \caption{AID}
    \begin{tabular}{l cc c}
    \toprule
    & Base & Novel & HM \\
    \midrule
    CLIP & 73.50	&70.40 &71.92 \\
    CoOp & 87.63	&70.37	&78.06\\
    Co-CoOp &92.63	&65.73	&76.89 \\
    MaPLe & 92.73	&74.57	&82.66 \\
    \midrule
    Ours (ViTAE) & \textbf{93.33}	&75.13	&83.25 \\
    \rowcolor{tabhighlight}
   Ours (ViT) &  92.90	&\textbf{76.03}	&\textbf{83.62} \\
    \bottomrule
    \end{tabular}
    \end{subtable}
    }
    ~
        \scalebox{0.8}{
    \begin{subtable}[t]{.32\textwidth}
    \centering
    \caption{RSICD}
    \begin{tabular}{l cc c}
    \toprule
    & Base & Novel & HM \\
    \midrule
    CLIP & 71.50   &60.20	&65.37 \\
    CoOp & 88.43	&60.20	&71.63 \\
    Co-CoOp & 92.37	&58.80	&71.86 \\
    MaPLe & 93.93	&56.27	&70.38 \\
    \midrule
    Ours (ViTAE) &94.93	&62.83	&75.61 \\
    \rowcolor{tabhighlight}
    Ours (ViT) &  \textbf{95.03}	&\textbf{64.64}	&\textbf{76.94} \\
    \bottomrule
    \end{tabular}
    \end{subtable}
    }
    ~
        \scalebox{0.8}{
    \begin{subtable}[t]{.32\textwidth}
    \centering
    \caption{UCM}
    \begin{tabular}{l cc c}
    \toprule
    & Base & Novel & HM \\
    \midrule
    CLIP & 80.60	&68.00	&73.77 \\
    CoOp & 93.60	&74.53	&82.98 \\
    Co-CoOp & 95.23	&71.57	&81.72 \\
    MaPLe & 97.70	&70.90	&82.17 \\
    \midrule
    Ours (ViTAE) & 97.00	&75.43	&84.87 \\
    \rowcolor{tabhighlight}
    Ours (ViT) &  \textbf{98.00}	&\textbf{80.00}	&\textbf{88.09}\\
    \bottomrule
    \end{tabular}
    \end{subtable}
    }
        \scalebox{0.8}{
    \begin{subtable}[t]{.32\textwidth}
    \centering
    \caption{WHURS19}
    \begin{tabular}{l cc c}
    \toprule
    & Base & Novel & HM \\
    \midrule
    CLIP & 73.10	&90.80  &80.99 \\
    CoOp & 95.20	 &82.40	&88.34 \\
    Co-CoOp & 97.10	&77.00	&85.89 \\
    MaPLe & 97.70	&88.03	&92.61 \\
    \midrule
    Ours (ViTAE) & \textbf{98.80}	 &91.10	&94.79 \\
    \rowcolor{tabhighlight}
    Ours (ViT)  &  98.77	&\textbf{93.70}	&\textbf{96.17} \\
    \bottomrule
    \end{tabular}
    \end{subtable}
    }
    ~
        \scalebox{0.8}{
    \begin{subtable}[t]{.32\textwidth}
    \centering
    \caption{NWPU}
    \begin{tabular}{l cc c}
    \toprule
    & Base & Novel & HM \\
    \midrule
    CLIP & 69.00	&63.00	&65.87 \\
    CoOp & 84.53	&66.97	&74.73 \\
    Co-CoOp & 89.27	&69.37	&78.07 \\
    MaPLe & 90.70	&72.70	&80.71\\
    \midrule
    Ours (ViTAE) & \textbf{90.97}	&\textbf{73.23}	&\textbf{81.14}\\
    \rowcolor{tabhighlight}
     Ours (ViT) &  90.80	& 72.80	&80.81 \\
    \bottomrule
    \end{tabular}
    \end{subtable}
    }

    \label{table1}
      \vspace{-0.5cm}
\end{table*}

\section{Experiments}
To assess the effectiveness of proposed method, we conducted extensive experiments using RSIs and medical images as examples, covering three distinct problem settings: 1) base-to-novel class generalization within a dataset, 2) cross-dataset transfer, and 3) domain generalization. Due to space limitations, more detailed special domain experiments are illustrated in supplementary materials. In this section, we offer a comprehensive overview of the utilized datasets and the evaluation metrics employed. Furthermore, we provide detailed insights into implementation specifics of our experiments. Subsequently, we conduct an in-depth analysis of our method's performance in each of aforementioned problem settings. Additionally, we performed ablation experiments to elucidate the effectiveness of our proposed approach.

\subsection{Datasets and Evaluation Metrics on RSIs}
The proposed method was evaluated on eight remote sensing datasets, namely MLRSNet~\cite{qi2020mlrsnet}, PatternNet~\cite{zhou2018patternnet}, RSSCN7~\cite{zou2015deep}, AID~\cite{xia2017aid}, RSICD~\cite{lu2017exploring}, UCM~\cite{yang2010bag}, WHURS19~\cite{Dai2011WHURS19}, and NWPU~\cite{cheng2017remote}.


Consistent with previous methods~\cite{khattak2023maple}, we employed accuracy and Harmonic Mean (HM) as evaluation metrics. The HM is computed as follows:

\begin{equation}
HM = \frac{{2 \times Acc_{base} \times Acc_{novel}}}{{Acc_{base} + Acc_{novel}}}
\end{equation}

Here, $Acc_{base}$ denotes the accuracy for base category, and $Acc_{novel}$ denotes the accuracy for novel category. It is critical to note that the reported results are averaged over three runs. For the base-to-novel generalization setting, experiments were conducted on all eight remote sensing datasets. In the cross-dataset generalization and domain generalization settings, MLRSNet was used as the source dataset, while the remaining datasets served as the target datasets.

\subsection{Implementation Details}
We implemented our method based on MaPLe and adopted similar training details. All experiments were conducted using a few-shot training strategy with 16 shots, randomly sampled for each class. We utilized the pre-trained ViT-B/16 CLIP model as the basis for prompt tuning. The training process for all models lasted for 5 epochs, employing a batch size of 4 and a learning rate of 0.0035. We utilized the SGD optimizer and performed the training on a single NVIDIA A100 GPU. The template for the word embeddings from CLIP is 'a photo of $category$'. We kept the hyperparameters consistent across all datasets to ensure fair comparisons. The language and visual control networks were implemented as two independent networks with the same architecture. Each network consisted of two linear layers followed by a ReLU activation layer. The architecture followed a Linear-ReLU-Linear pattern. By adhering to these training details and network configurations, we aimed to establish a solid foundation for comparison and replication of the experiments, maintaining consistency with the MaPLe framework.

\begin{table*}[!t]
\small
\centering
    \caption{Comparisons between our method with state-of-the-art methods for cross-dataset generalization with MLRSNet dataset as the
source domain and remaining remote sensing datasets as the target domains. The best results are shown in bold.
    }
     \vspace{-0.35cm}
	\scalebox{0.70}{
\begin{tabular}{l c cccccccc}
    \toprule
    & \textbf{\ \ Source\ \ } & \multicolumn{8}{c}{\textbf{Target}} \\
    \cmidrule{2-2} \cmidrule(l){3-10}
   & MLRSNet & PatternNet & RSSCN7 & AID & RSICD & UCM & WHURS19 & NWPU & Average\\
    \midrule
    \ \  CoOp & 72.53	&66.97	&69.03	&67.30	&63.50	&77.57	&85.47	&70.43	&71.60\\
    \ \ Co-CoOp \  & 71.70	&65.67	&68.80	&66.63	&62.57	&76.40	&85.33	&70.30	&70.92\\
    \ \ MaPLe & 76.83	&\textbf{68.53}	&71.43	&65.13	&59.53	&\textbf{79.90}	&85.23	&72.80	&72.42  \\
    \midrule
    \rowcolor{tabhighlight} \ \ Ours & \textbf{77.73}	&67.13	&\textbf{71.60}	&\textbf{68.73}	&\textbf{64.17}	&78.50	&\textbf{86.97}	&\textbf{72.87}	&\textbf{73.46}  \\
    \bottomrule
    \end{tabular}
    }

    \label{table2}
  \vspace{-0.4cm}
\end{table*}

\begin{table*}[!t]
\small
\centering
    \caption{Comparisons between our method with SOTA methods for single-source multi-target domain generalization with MLRSNet dataset as the source domain and remaining datasets as the target domains. The best results are shown in bold.
    }
     	  \vspace{-0.35cm}
	\scalebox{0.70}{
\begin{tabular}{l c cccccccc}
    \toprule
    & \textbf{\ \ Source\ \ } & \multicolumn{8}{c}{\textbf{Target}} \\
    \cmidrule{2-2} \cmidrule(l){3-10}
   & MLRSNet & PatternNetv2 & RSSCN7v2 & AIDv2 & RSICDv2 & UCMv2 & WHURS19v2 & NWPUv2 & Average\\
    \midrule
    \ \  CoOp & 72.53	&66.97	&69.07	&67.13	&64.27	&77.40	&85.20	&71.17	&71.72\\
    \ \ Co-CoOp \  & 71.70	&65.57	&69.37	&67.13	&62.73	&75.70	&84.83	&70.97	&71.00\\
    \ \ MaPLe & 76.83	&68.03	&\textbf{72.50}	&64.90	&59.73	&\textbf{78.93}	&83.07	&73.17	&72.15  \\
    \midrule
    \rowcolor{tabhighlight} \ \ Ours & \textbf{77.73}	&\textbf{68.27}	&72.10	&\textbf{68.33}	&\textbf{64.57}	&77.30	&\textbf{85.80}	&\textbf{73.37}	&\textbf{73.43}  \\
    \bottomrule
    \end{tabular}
    }

    \label{table3}
  \vspace{-0.4cm}
\end{table*}

\subsection{Generalization from Base-to-Novel Classes}
The primary objective of prompt learning is to facilitate the transfer of large-scale models to downstream tasks, with the challenge of ensuring effective generalization from base to novel classes. To address this critical issue, we conducted comprehensive experiments and comparisons on all eight remote sensing recognition datasets. Leveraging the capabilities of LRSM, we employed two distinct pre-trained models, namely ViT and ViTAE, to incorporate essential domain-specific knowledge for remote sensing applications. Our method was evaluated against state-of-the-art techniques such as zero-shot CLIP, as well as prompt learning methods including CoOp, CoCoOp, and MaPLe, establishing a robust benchmark for performance assessment. The detailed results and comparative analyses are presented in Table~\ref{table1}.

In comparison to the leading MaPLe approach, our proposed method exhibited significant performance improvements across both the base and novel categories in all eight datasets. Particularly noteworthy was the superior overall performance achieved by our ViT-based approach. For the base categories, our method showed a noteworthy gain, elevating performance from 93.07$\%$ to 93.7$\%$. Notably, the improvements were even more substantial for novel categories, where performance skyrocketed from 73.79$\%$ to 75.81$\%$. Taking into account the performance across both base and novel classes, our method outperformed MaPLe by an average absolute gain of 2.94$\%$. Of particular interest was the exceptional performance boost observed on the RSICD dataset, where our method achieved an impressive gain of 6.56$\%$. However, it is worth mentioning that all prompt learning methods failed to match the performance of zero-shot CLIP in the novel categories of the MLRSNet dataset, indicating that the conventional word template may not be well-suited for prompt learning in this specific context.

An intriguing observation emerged when comparing the ViTAE model, which possesses a deeper architecture and greater expressive capacity, with the commonly leveraged ViT model. Surprisingly, while the overall performance based on ViTAE surpassed that of MaPLe, it fell slightly short when compared to the ViT-based approach. This intriguing phenomenon suggests the existence of an upper limit to the utilization of remote sensing foundation knowledge, where a deeper architecture may not always be the optimal choice for prompt learning. Furthermore, a detailed analysis revealed that our ViTAE-based method delivered enhanced performance for the base categories in the AID, WHURS19, and NWPU datasets. However, a relatively lower performance was observed in the novel categories, consistently aligning with our earlier findings and emphasizing the intricate relationship between model depth, prompt learning, and dataset characteristics.

\begin{table}[t!]
\small
\tablestyle{6pt}{0}
\addtolength{\tabcolsep}{-6pt}
    \tabstyle{1.0pt}
    \setlength{\tabcolsep}{3.2pt}
    \caption{Comparison with existing methods in base-to-novel generalization on medical image classification datasets. The best results are shown in bold.
    }
     \vspace{-0.4cm}
    \scalebox{1.0}{
    \begin{subtable}[t]{.25\textwidth}
    \centering
    \caption{\textbf{Average over datasets}}
        \begin{tabular}{l cc c}
    \toprule
    & Base & Novel & HM \\
    \midrule
    CLIP &49.83	&41.83	&45.18 \\
    CoOp & 51.59	&43.77	&46.81 \\
    Co-CoOp & 64.45	&43.16	&49.45 \\
    MaPLe & 62.39	&44.40 &49.01 \\
    \midrule
    \rowcolor{tabhighlight}
       Ours  &  \textbf{66.11}	&\textbf{48.75}	&\textbf{53.08}\\
        &\textcolor{red}{+1.66}	&\textcolor{red}{+4.35}	&\textcolor{red}{+3.63}\\
    \bottomrule
    \end{tabular}

    \end{subtable}
    }
        ~
    \begin{subtable}[t]{.20\textwidth}
    \centering
    \caption{All datasets}
\includegraphics[width=0.8\linewidth]{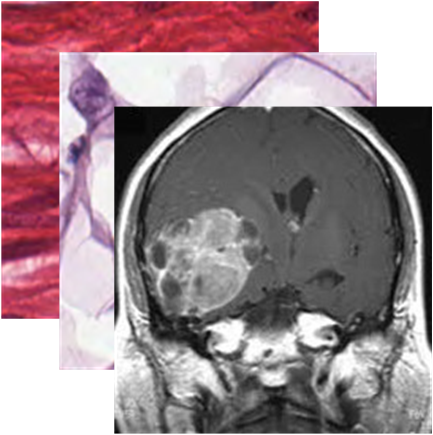}
    \end{subtable}
        \label{table4}
      \vspace{-0.6cm}
\end{table}
\subsection{Cross-dataset Evaluation}
In order to showcase the proposed method's ability to generalize across datasets, we employed the MLRSNet to conduct domain-controlled prompt learning and directly evaluated the model on the remaining seven datasets. The comparative results between our method and other popular algorithms are presented in Table~\ref{table2}. Remarkably, our method achieved superior performance on the MLRSNet, resulting in a substantial performance improvement of nearly 1$\%$. Furthermore, the most significant performance boost was observed in the RSICD dataset, indicating that the domain-controlled prompt learning approach is particularly well-suited for the RSICD dataset. Although our method did not yield favorable results on the PatternNet and UCM datasets, it surpassed all existing methods in terms of overall performance, with a noteworthy improvement of 1.04$\%$. These findings demonstrate the effectiveness of our method in terms of cross-dataset generalization. 

\subsection{Domain Generalization}
To further validate the generalization ability of our proposed method, we conducted an evaluation in the domain generalization setting, adhering to the experimental protocol employed by prior studies. Our approach was compared against other state-of-the-art algorithms, and the comparative results are presented in Table~\ref{table3}. Remarkably, our method consistently outperforms the competing algorithms, achieving the highest average performance with a noteworthy 1.28$\%$ improvement. It is important to note that while our method may encounter challenges when applied to the RSSCN7v2 and UCMv2 datasets, it excels on the RSICDv2 dataset, showcasing an impressive performance gain of 4.84$\%$. These findings underscore the efficacy of incorporating domain-controlled prompt learning in enhancing the generalization and robustness of visual-linguistic models like CLIP for the analysis of remote sensing images.

\subsection{Experiments on Other Domain}
To further validate the effectiveness of our proposed method, we conducted comprehensive experiments on medical domain datasets, including BTMRI~\cite{BTMRI}, CHMNIST~\cite{CHMNIST}, and CCBTM~\cite{CCBTM}. The comparative results between our method and other advanced algorithms are summarized in Table~\ref{table4} (Accuracy and HM as metrics).  Specifically, our method achieves an impressive 1.66$\%$ performance improvement for base categories and an even more substantial 4.35$\%$ improvement for  novel categories. When considering the overall performance metric, Harmonic Mean (HM), our method exhibits a significant 3.63$\%$ improvement compared to other algorithms. These compelling results indicate the robustness and efficacy of our proposed approach in medical domain datasets. Due to the space limitation, we provide more detailed experimental results and analysis in the supplementary material.

\begin{table}[t]
	\centering
	\scriptsize
	\footnotesize
	\renewcommand{\tabcolsep}{4.0mm}
	\caption{ Ablation study of domain-controlled prompt learning in different branches. VC and LC individually denote Visual and Language domain-controlled prompt learning.}
 	  \vspace{-0.35cm}
	\scalebox{0.75}{
		\begin{tabular}{c|ccc}
			\hline
			\multicolumn{1}{c|}{Methods}  &\multicolumn{1}{c}{Base}  &\multicolumn{1}{c}{Novel}  &\multicolumn{1}{c}{HM} \\
			\hline
   		\multicolumn{1}{c|}{Baseline} &97.70	&70.90	&82.17 \\
            \multicolumn{1}{c|}{Baseline+VC}  &97.80	&76.43	&85.80\\
			\multicolumn{1}{c|}{Baseline+LC}   &97.60	&73.33	&83.74\\
   		\multicolumn{1}{c|}{Ours}   &\textbf{98.00}	&\textbf{80.00}	&\textbf{88.09}\\
			\hline	
	\end{tabular} }
	\label{table5}
   \vspace{-0.3cm}
\end{table}

\subsection{Ablation Study}

\textbf{Domain-Controlled Prompt Learning.} In order to analyze the impact of different components in domain-controlled prompt learning, we conducted separate experiments for both the visual and language branches. The evaluations were performed on the UCM datasets, and the results are summarized in Table~\ref{table5}. It is evident that incorporating domain-controlled prompt learning in both branches leads to performance improvements. Specifically, controlling the visual branch yields substantial performance gains, particularly in the case of novel categories, resulting in an overall improvement of 3.63$\%$. On the other hand, domain-controlled prompt learning in the language branch contributes to a relatively lower performance boost but still achieves an overall improvement of 1.57$\%$. These findings highlight the effectiveness of domain-controlled prompt learning in benefiting both the visual and language branches, ultimately enhancing the accuracy of remote sensing image recognition.

\begin{table}[t]
	\centering
	\scriptsize
	\footnotesize
	\renewcommand{\tabcolsep}{5.0mm}
	\caption{ Ablation study of overfitting-tackling strategies.}
 	  \vspace{-0.35cm}
	\scalebox{0.75}{
		\begin{tabular}{c|ccc}
			\hline
			\multicolumn{1}{c|}{Methods}  &\multicolumn{1}{c}{Base}  &\multicolumn{1}{c}{Novel}  &\multicolumn{1}{c}{HM} \\
			\hline
      	\multicolumn{1}{c|}{Baseline} &97.70	&70.90	&82.17 \\
   		\multicolumn{1}{c|}{Dropout(0.3)} &97.78	&77.83	&86.67 \\
            \multicolumn{1}{c|}{Dropout(0.5)}  &97.30	&77.67	&86.38\\
            \multicolumn{1}{c|}{Mutation(0.05)}  &97.60	&71.67	&82.65\\
            \multicolumn{1}{c|}{Mutation(0.1)}  &97.20	&71.57	&82.44\\
   		\multicolumn{1}{c|}{Ours}   &\textbf{98.00}	&\textbf{80.00}	&\textbf{88.09}\\
			\hline	
	\end{tabular} }
	\label{table6}
   \vspace{-0.4cm}
\end{table}

\textbf{Different Overfitting-Tackling Strategies.} In our method, we adopt the proposed noisy-adding strategy to explicitly solve the overfitting problem. As mentioned before, adopting dropout or mutation operations seems to be a plausible solution. Thus, we implement a series of experiments on the UCM dataset to distinguish our method from other strategies, and the experimental results are shown in Table~\ref{table6}. The dropout and mutation operations could both bring overall performance improvements since handling the overfitting problem. The dropout with a rate of 0.3 has a better performance than the dropout with a rate of 0.5, and the mutation with 5 percent has a better performance  than the mutation with 10 percent.  Though these operations could bring some performance improvements, our proposed noisy-adding strategy could have obviously better performance improvements.  This phenomenon suggests the local sampling in dropout and point jittering in mutation are insufficient in escaping suboptimal solutions, yet our method helps the network have a broader solution exploration in a global domain oscillation manner.

 \begin{figure}[t]
	\begin{center}
		\includegraphics[width=0.83\linewidth]{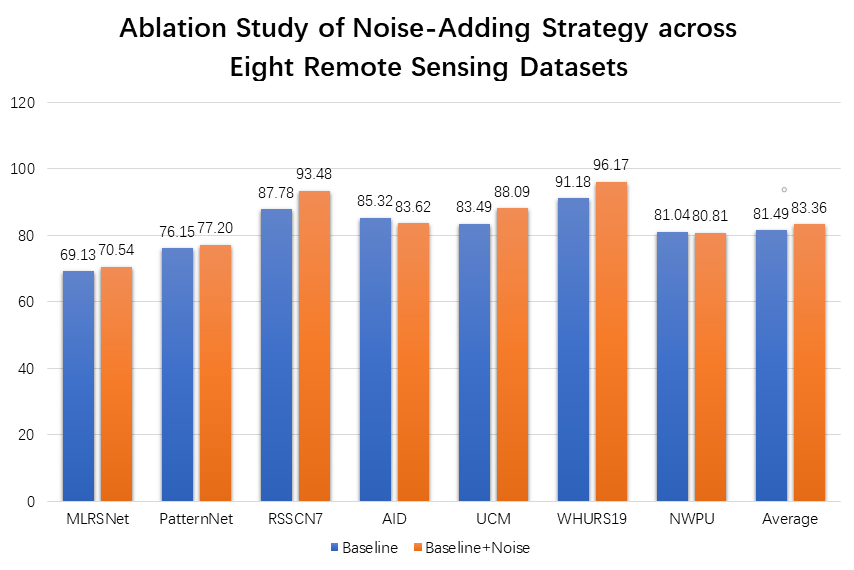}
	\end{center}
  \vspace{-0.4cm}
	\caption{The ablation study of noise-adding strategy across eight remote sensing datasets. }
	\label{fig3}
 \vspace{-0.8cm}
\end{figure}

\textbf{Noise-Adding Strategy across Datasets.} 
To comprehensively assess the impact of the noise-adding strategy, we conducted experiments across eight diverse remote sensing datasets. The performance gains achieved by incorporating the noise-adding strategy are illustrated in Figure~\ref{fig3}. The results demonstrate that the noise-adding strategy consistently improves performance across the majority of datasets, with only minor performance decreases observed in the NWPU and AID datasets. Remarkably, the noise-adding strategy leads to an overall performance improvement of 1.87$\%$. This observation highlights the effectiveness of the proposed strategy as a generalized approach to mitigate overfitting, thereby boosting performance.

\section{Conclusion}
Focusing on the neglected natural-to-specific adaptation challenge, we introduce large-scale specific domain foundation models to provide specific domain knowledge and further perform domain-controlled prompt learning in both visual and language branches for specific domain images. To overcome the base-to-novel overfitting challenge, a novel noisy adding strategy is proposed to explicitly escape the suboptimal solutions in a global domain oscillation manner. To validate the effectiveness of our method, we conduct extensive experiments using specific domain datasets like RSIs and medical images to demonstrate the superiority of our proposed approach.

\bibliography{aaai24}
\clearpage

\appendix
\begin{center}
\huge Supplemental Materials for  
\par
~\\
\Large Domain-Controlled Prompt Learning
\end{center}

\begin{table*}[t!]
\small
\tablestyle{6pt}{0}
\addtolength{\tabcolsep}{-6pt}
    \tabstyle{1.0pt}
    \setlength{\tabcolsep}{3.2pt}
    \caption{Comparison with existing methods in base-to-novel generalization on medical image classification datasets. The best results are shown in bold. \textcolor{red}{Red:} performance gain, \textcolor{blue}{Blue:} performance drop.
    }
    \scalebox{1.0}{
    \begin{subtable}[t]{.25\textwidth}
    \centering
    \caption{\textbf{Average over datasets}}
        \begin{tabular}{l cc c}
    \toprule
    & Base & Novel & HM \\
    \midrule
    CLIP &49.83	&41.83	&45.18 \\
    CoOp & 51.59	&43.77	&46.81 \\
    Co-CoOp & 64.45	&43.16	&49.45 \\
    MaPLe & 62.39	&44.40 &49.01 \\
    \midrule
    \rowcolor{tabhighlight}
       Ours  &  \textbf{66.11}	&\textbf{48.75}	&\textbf{53.08}\\
        &\textcolor{red}{+1.66}	&\textcolor{red}{+4.35}	&\textcolor{red}{+3.63}\\
    \bottomrule
    \end{tabular}

    \end{subtable}
    }
        ~
    \begin{subtable}[t]{.20\textwidth}
    \centering
    \caption{All datasets}
\includegraphics[width=0.8\linewidth]{all.png}
    \end{subtable}
\scalebox{1.0}{
    \begin{subtable}[t]{.25\textwidth}
    \centering
    \caption{BTMRI}
    \begin{tabular}{l cc c}
    \toprule
    & Base & Novel & HM \\
    \midrule
    CLIP &50.60	&51.20	&50.89 \\
    CoOp & 48.93	&53.30	&51.02 \\
    Co-CoOp & 52.37	&52.80	&52.58 \\
    MaPLe & 53.67	&61.60 &57.36 \\
    \midrule
    \rowcolor{tabhighlight}
       Ours  &  \textbf{57.00}	&\textbf{67.13}	&\textbf{61.65}\\
       &\textcolor{red}{+3.33}	&\textcolor{red}{+5.53}&\textcolor{red}{+4.29}\\
    \bottomrule
    \end{tabular}
    \end{subtable}
    }
            ~
    \begin{subtable}[t]{.20\textwidth}
    \centering
    \caption{BTMRI dataset}
\includegraphics[width=0.8\linewidth]{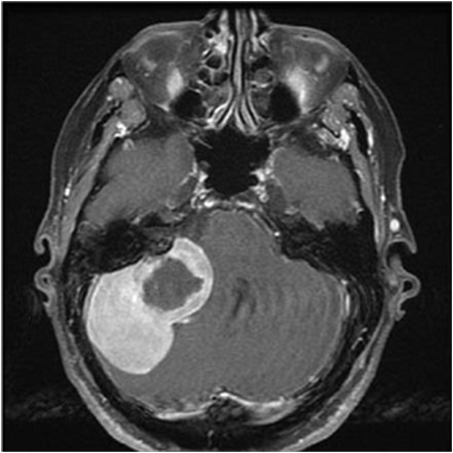}
    \end{subtable}
         ~
        \scalebox{1.0}{
    \begin{subtable}[t]{.25\textwidth}
    \centering
    \caption{CHMNIST}
    \begin{tabular}{l cc c}
    \toprule
    & Base & Novel & HM \\
    \midrule
    CLIP & 31.60 & \textbf{27.40} & 29.35 \\
    CoOp & 41.70 & 25.67 & 31.78 \\
    Co-CoOp & 74.30 & 25.30 & 37.74 \\
    MaPLe & 74.03 & 25.10 & 37.49 \\
    \midrule
    \rowcolor{tabhighlight}
     Ours  & \textbf{74.83} &	25.85	&\textbf{38.43}
 \\
  &  \textcolor{red}{+0.50} &\textcolor{blue}{-1.55}	&\textcolor{red}{+0.69} \\
    \bottomrule
    \end{tabular}
    \end{subtable}
    }
    ~
    \begin{subtable}[t]{.20\textwidth}
    \centering
    \caption{CHMNIST dataset}
\includegraphics[width=0.8\linewidth]{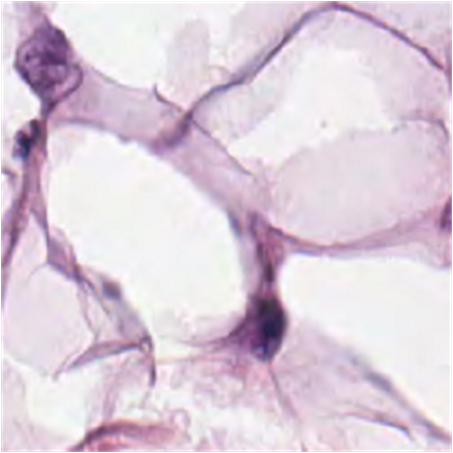}
    \end{subtable}
    ~
        \scalebox{1.0}{
    \begin{subtable}[t]{.25\textwidth}
    \centering
    \caption{CCBTM}
    \begin{tabular}{l cc c}
    \toprule
    & Base & Novel & HM \\
    \midrule
    CLIP &\textbf{ 67.30} & 46.90 & 55.28 \\
    CoOp & 64.13  &	52.33 & 57.63 \\
    Co-CoOp & 66.67	&51.37	&58.03 \\
    MaPLe & 59.47	&46.50	&52.19 \\
    \midrule
    \rowcolor{tabhighlight}
    Ours  &  66.50	&\textbf{53.27}	&\textbf{59.15}\\
    & \textcolor{blue}{-0.80}	&\textcolor{red}{+0.94} &\textcolor{red}{+1.52}\\
    \bottomrule
    \end{tabular}
    \end{subtable}
        }
        ~
    \begin{subtable}[t]{.20\textwidth}
    \centering
    \caption{CCBTM dataset}
\includegraphics[width=0.8\linewidth]{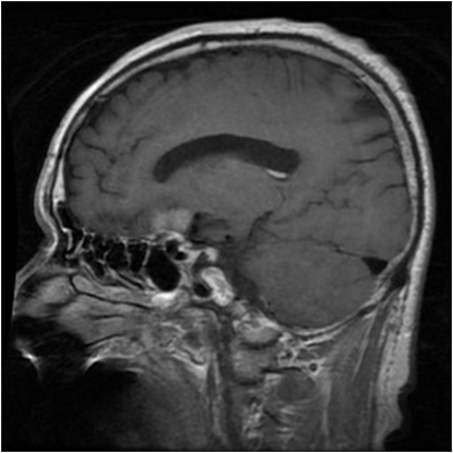}
    \end{subtable}
    \label{table7}
\end{table*}

\section{Experiments on Medical Images}
To assess the effectiveness of the proposed method, we conducted extensive experiments on medical images as other examples. The experiments also cover three distinct problem settings: 1) base-to-novel class generalization within a dataset, 2) cross-dataset transfer, and 3) domain generalization. In this section, we detailedly introduce the leveraged LSDM for medical images and utilized medical image recognition datasets.  Subsequently, we conduct an in-depth analysis of our method's performance in each of the aforementioned problem settings. The  compelling experimental results demonstrate the superiority of our proposed approach.

\subsection{LSDM and Datasets}
As mentioned before, LSDM aims at providing specific domain knowledge to achieve domain-controlled prompt learning.  Harnessing the power of a meticulously curated dataset comprising over one million images, MedSAM~\cite{ma2023segment}, the inaugural foundation model designed for universal
medical image segmentation, not only outperforms existing state-of-the-art segmentation foundation models, but also exhibits comparable or even superior performance to specialist models. Inspired by this success, we adopt the newly introduced MedSAM~\cite{ma2023segment} as the ideal foundation model for our medical image recognition tasks. We adopt three medical image recognition datasets for experiments: the Brain Tumor MRI Dataset (BTMRI)~\cite{BTMRI}, Colorectal Histology MNIST (CHMNIST)~\cite{CHMNIST}, and Crystal Clean: Brain Tumors MRI Dataset (CCBTM)~\cite{CCBTM}. 
To ensure consistency and comparability with previous studies, we employ accuracy and Harmonic Mean (HM) as the evaluation metrics for our experiments.  Meanwhile, the same experimental details as mentioned before are adopted  for medical images. 

\subsection{Generalization from Base-to-Novel Classes}
In our evaluation, we compared the proposed method against several state-of-the-art techniques, including zero-shot CLIP, and prompt learning methods like CoOp, CoCoOp, and MaPLe, thus establishing a robust benchmark for performance assessment. As depicted in Table~\ref{table7}, our method outperforms all others in all the medical image recognition datasets. For the base categories, our method demonstrates a remarkable 1.66$\%$ performance improvement. Additionally, when it comes to the novel categories, our method achieves an even more impressive 4.35$\%$ performance improvement. In terms of the harmonic mean, our method elevates the performance from the second-best at 49.45$\%$ to an outstanding 53.08$\%$. Specifically, our approach excels in the BTMRI dataset, achieving over 3$\%$ performance improvement across all categories. Notably, it achieves a remarkable 5.53$\%$ performance improvement specifically for the novel categories. While our method may not achieve the top performance in terms of novel categories in the CHMNIST dataset, it does obtain the best performance among all prompt-learning methods. Moreover, our method continues to dominate when considering the overall performance evaluation metric, HM. Similar encouraging results are observed in the CCBTM dataset. For the base categories, our method obtains the best performance compared to all prompt-learning methods. Furthermore, the harmonic mean shows a 1.52$\%$ performance gain, further highlighting the strength and effectiveness of our proposed approach.

\begin{table}[!t]
\small
\centering
    \caption{Comparisons between our method with SOTA methods for single-source multi-target domain generalization with BTMRI dataset as the source domain and remaining datasets as the target domains. The best results are shown in bold. \textcolor{red}{Red:} performance gain, \textcolor{blue}{Blue:} performance drop.
    }
	\scalebox{0.9}{
\begin{tabular}{l c ccc}
    \toprule
    & \textbf{\ \ Source\ \ } & \multicolumn{3}{c}{\textbf{Target}} \\
    \cmidrule{2-2} \cmidrule(l){3-5}
   & BTMRI & CHMNIST & CCBTM &Average\\
    \midrule
    \ \  CoOp &39.00	&15.43	&30.97	&28.47	\\
    \ \ Co-CoOp \  & 52.93	&16.57	&29.13	&32.88	\\
    \ \ MaPLe & 57.63	&16.53	&33.13	&35.76	  \\
    \midrule
    \rowcolor{tabhighlight} \ \ Ours & \textbf{62.00}	&\textbf{17.43}	&\textbf{33.31}	&\textbf{37.58}	 \\
        & \textcolor{red}{+4.37}	&\textcolor{red}{+0.86} &\textcolor{red}{+0.18}&\textcolor{red}{+1.82}\\
    \bottomrule
    \end{tabular}
    }

    \label{table8}
\end{table}

\subsection{Cross-dataset Evaluation}
To demonstrate the generalization capability of our proposed method across datasets, we conducted domain-controlled prompt learning using the BTMRI dataset and then directly evaluated the trained model on the remaining medical image datasets. The comparative results are presented in Table~\ref{table8}. Clearly, our method exhibits superior performance in the BTMRI dataset, achieving an impressive 4.37$\%$ performance improvement. Moreover, this improvement extends to the CHMNIST dataset, where we observe a 0.86$\%$ increase in performance. Although the improvement for the CCBTM dataset is relatively small, at 0.18$\%$, the average accuracy across the datasets experiences a notable boost of 37.58 and achieves a 1.82$\%$ improvement. These results validate the effectiveness of our proposed approach in generalizing across diverse medical image datasets. The performance improvements in different datasets demonstrate the robustness and adaptability of our method, further supporting its potential for broader applicability in medical image recognition tasks.

\subsection{Domain Generalization}
To further assess the generalization ability of our proposed method, we conducted an evaluation in the domain generalization setting, using the BTMRI dataset as the source dataset. As shown in Table~\ref{table9}, the results from this evaluation also demonstrate the effectiveness of our approach. Specifically, in the BTMRI dataset, our method achieves a notable 4.37$\%$ performance improvement, showcasing its capability to adapt and perform well even in the domain it was trained on. Additionally, when tested on the CHMNISTv2 dataset, our method shows a commendable 1.20$\%$ performance improvement, further supporting its generalization capabilities across different datasets. While our method may not outperform the previous state-of-the-art method, MaPLe, in the CCBTMv2 dataset, it still achieves a competitive 37.68$\%$ accuracy and a respectable 1.70$\%$ performance improvement. These results highlight the robustness of our proposed approach and its capacity to deliver meaningful performance gains even in challenging domain generalization scenarios. Overall, these findings reinforce the superior performance and generalization abilities of our method.

\begin{table}[!t]
\small
\centering
    \caption{Comparisons between our method with SOTA methods for cross-dataset generalization with BTMRI dataset as the
source domain and remaining remote sensing datasets as the target domains. The best results are shown in bold. \textcolor{red}{Red:} performance gain, \textcolor{blue}{Blue:} performance drop.
    }
	\scalebox{0.9}{
\begin{tabular}{l c ccc}
    \toprule
    & \textbf{\ \ Source\ \ } & \multicolumn{3}{c}{\textbf{Target}} \\
    \cmidrule{2-2} \cmidrule(l){3-5}
   & BTMRI & CHMNISTv2 & CCBTMv2 &Average\\
    \midrule
    \ \  CoOp & 39.00	&15.53	&31.10	&28.54	\\
    \ \ Co-CoOp \  & 52.93	&16.03	&29.10	&32.69	\\
    \ \ MaPLe & 57.63	& 15.93	&34.40	&35.98	  \\
    \midrule
    \rowcolor{tabhighlight} \ \ Ours & \textbf{62.00}	&\textbf{17.23}	&\textbf{33.81}	&\textbf{37.68}	 
    \\
        & \textcolor{red}{+4.37}	&\textcolor{red}{+1.20} &\textcolor{blue}{-0.59}&\textcolor{red}{+1.70}\\
    \bottomrule
    \end{tabular}
    }

    \label{table9}
\end{table}

\end{document}